# *Les outils numériques en santé au travail, freins ou leviers de construction des dynamiques pluridisciplinaires ?*

Cédric GOUVENELLE[1, 2], Flora MAUDHUY[1], Florence THORIN[1]

[1]APST18, c.gouvenelle@apst18.com ; f.maudhuy@apst18.com
[2]Université Clermont Auvergne, CNRS, Acté, F-63000 Clermont–Ferrand, France

**RÉSUMÉ**

L'arrivée des plateformes numériques a révolutionné la santé au travail en donnant la possibilité aux Services de Prévention et de Santé au Travail Interentreprises (SPSTI) de se doter de bases de données permettant d'offrir aux professionnels de nouvelles possibilités d'agir. Cependant, dans un secteur d'activité qui se pose la question du développement de la pluridisciplinarité depuis 20 ans, l'arrivée de nouveaux outils peut parfois sembler être rapidement une solution. L'étude, menée dans un SPSTI précurseur en termes de développement des outils numériques, a pour objectif de faire le point sur les modalités et les impacts des transformations instrumentales et organisationnelles pour les professionnels de santé comme pour les membres des équipes techniques du SPSTI. Il s'agit de mettre en lumière les freins et les leviers ainsi que les différentes possibilités d'accompagnement à envisager.

**MOTS-CLÉS**

Santé au Travail, Data, pluridisciplinarité, Activité, Asymétries

## 1   LA SANTE AU TRAVAIL, UN SECTEUR EN MUTATION

La Loi santé travail apporte de nouvelles prescriptions aux Services de Prévention en Santé au Travail Interprofessionnels (SPSTI), notamment en termes d'offres, de traçabilité, d'évaluation des activités et de sécurisation des données. Face à ces transformations, de nouveaux outils et de nouvelles organisations naissent afin d'atteindre une ambition de 2002 : construire un fonctionnement qui allie pluridisciplinarité et performance des équipes pluriprofessionnelles. L'APST18 a cherché à anticiper les transformations légales et sociétales futures en engageant dès 2017 une réflexion prospective avec des start-up innovantes sur les moyens à mettre en œuvre afin de répondre au mieux aux besoins futurs des adhérents comme des travailleurs par un processus de co-construction avec les personnes de tout statut du SPSTI. A la date de l'étude, celui-ci est doté de 14 équipes autonomes, composées chacune d'un médecin, d'un infirmier santé travail (IDEST), d'un assistant de l'équipe pluriprofessionnelle, d'1/3 de technicien HSE et 1/3 d'ergonome pour un effectif dédié.

## 2   CADRE THEORIQUE

Les travaux s'appuient sur une approche centrée sur l'activité des personnes, leur perception, leur sensibilité (Coutarel, 2013; Récopé et al., 2013). La data est étudiée comme un outil de travail des personnes. Les personnes ne sont pas au services des outils mais coconstructeurs. Ils sont acteurs de la création de leur propre réalité, par la construction collective, les apprentissages, ils sont en position de modeler les outils et les organisations. Des freins ont été documentés au sein des SPSTI (Barlet, 2015; Caroly, 2013; Caroly et al., 2011; Caroly & Sallah, 2015). Caroly (2013) montre que la distribution même des tâches introduit au sein des SSTI des distorsions statutaires entre les métiers qui n'ont pas tous la même visibilité. Certains ont des positions particulières qui les mettent dans une situation de dépendance au médecin du travail qui apparaît en position asymétrique favorable et oriente l'action des autres membres de l'équipe pluriprofessionnelle. Cette dépendance est issue des prescriptions institutionnelles et organisationnelles, mais aussi de la reconnaissance du statut, de l'expertise comme de sa reconnaissance sociale (Bourdieu, 1979). Elle tient aussi d'un manque d'homogénéité dans la construction de règles de métier dans les métiers de l'équipes pluriprofessionnelle. Les possibilités de déploiement des débats interprofessionnels apparaissent faibles (Barlet, 2015; Caroly & Sallah, 2015).

Pour autant, certaines circonstances encouragent l'engagement des personnes en position asymétrique défavorable, le développement du débat et au final de la pluridisciplinarité (Gouvenelle, 2021). L'acceptabilité technique de la data, la perception de son utilité ou de son utilisabilité émerge comme central pour les utilisateurs (Nielsen, 1994). Il ne s'agit pas seulement de la prise en main de l'outil mais qu'il réponde efficacement à un besoin identifié des utilisateurs. L'outil doit être perçu comme un facilitateur producteurs de dynamiques et non comme des frein au développement de l'activité (Barcellini et al., 2014).

## 3 METHODOLOGIE

La méthodologie est porteuse d'éléments de comparaison qui peuvent faire émerger des régularités dans les matériaux. Ces régularités s'entendent en termes de perception et de conduite des personnes, afin de pouvoir situer les régulations face aux modalités d'utilisation des outils ou organisationnelles mais aussi de constructions de normes collectives de métier (Garrigou et al., 2022).

La méthodologie de l'étude est construite en 4 phases. Des matériaux sont construits entre 2017 et 2021 lors d'une recherche dont l'objectif est de comprendre les dynamiques sous-jacentes aux prises de rôles (Gouvenelle, 2021). Elle a permis de déterminer des caractéristiques d'un fonctionnement pluridisciplinaire des équipes pluriprofessionnelles. Pour cela, il a été réalisé de :

- L'analyse de l'activité des personnes
- L'objectivation de l'activité : catégorisation des rôles, des actions réalisées…
- La co-construction des organisations et des instruments (projets et groupes de travail sur les transformations des organisations et sur la conception/évolutions des outils)

Entre 2020 et 2023, une évaluation de la perception du vécu des acteurs (travailleurs et employeurs) est réalisée par le biais de la plateforme interconnectée. Elle concerne également les membres du SPSTI. La réalisation s'appuie sur :

- o 475 questionnaires de satisfaction pour les travailleurs
- o 100 questionnaires de satisfaction employeurs
- o 48 Questionnaires sur l'utilisation de la data et de l'outil pour les professionnels du SPSTI
- o 40 Entretiens réalisés

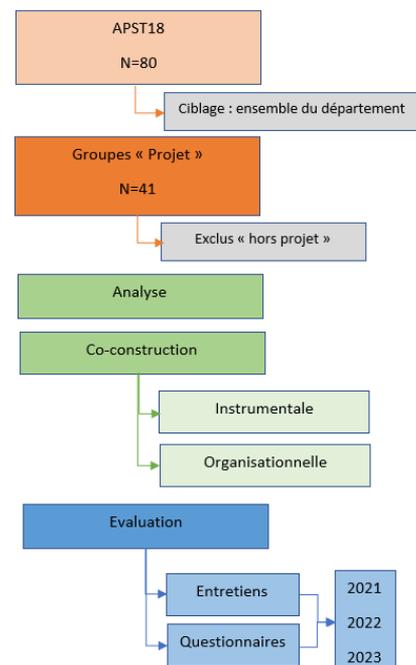

*Figure 1 Flow chart Méthodologie*

## 4 RESULTATS

### 4.1 La perception de l'introduction de la data

Les résultats de 3 ans d'études des adhérents et des travailleurs font émerger la satisfaction de l'introduction de la plateforme interconnectée en santé travail dans les processus de suivi et d'accompagnement des travailleurs et des entreprises. Cette satisfaction est différenciée en fonction des populations. Les travailleurs notent une transformation du rapport au professionnel de santé lors de la visite. 76% des travailleurs interrogés estiment que la visite pré-connectée permet aux professionnels de santé de mieux connaître leur état de santé et leurs expositions (Figure 2).

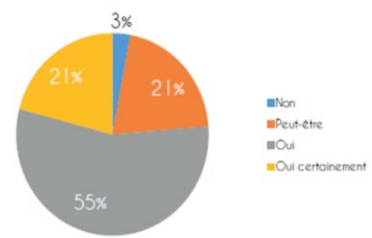

*Figure 2 Perception de la visite préconnectée par les travailleurs*

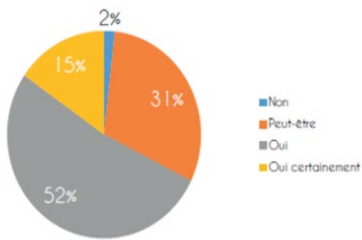

*Figure 3 Perception de la visite préconnectée par les employeurs*

Celui-ci apparaît plus en capacité d'orienter et de conseiller les travailleurs du fait d'une meilleure connaissance des métiers, de leurs expositions et risques spécifiques. Le professionnel de santé, en fonction des expositions auxquelles est soumis le travailleur, mais aussi de l'entretien, prescrit des examens complémentaires, oriente vers des spécialistes ou initie une intervention en entreprise. Une réponse aux besoins des travailleurs satisfait également les employeurs. Ils sont 67% à estimer que cette visite pré-connectée permet à l'équipe pluriprofessionnelle de mieux prendre en charge la santé de l'ensemble des collaborateurs (Figure 3). Cette reconnaissance de la qualité du service du SPSTI est nouvelle là où habituellement émerge principalement la visite médicale.

Les professionnels signifient eux-aussi, (83%), que l'utilisation quotidienne des outils numériques dans leur activité de travail est jugée satisfaisante (Figure 4). Les outils, dans le discours des médecins du travail, émergent comme une aide technique et cognitive : *« Le fait d'avoir les expositions, ça donne des informations auxquelles tu ne penserais pas […] quand on ne connait pas les entreprises »* ou *« Il y a des sujets qui ressortent qui ne ressortaient pas avant. C'est plus simple à aborder aujourd'hui car c'est le travailleur qui a préalablement abordé le sujet. ça devient plus simple à requestionner en repartant de ce que le travailleur a répondu»*. La data et les suggestions issues de l'IA sont perçus comme comme une aide à la décision pas une dépossession professionnelle. Les données à disposition permettent aussi une meilleure connaissance de l'activité, de l'entreprise et des expositions des travailleurs ce qui contribue pour les équipes pluriprofessionnelles à mieux conseiller l'employeur, adapter le suivi et les ressources et coconstruire en équipe les méthodologies d'intervention supportant les actions en milieu de travail (AMT). Les données sont issues :

- Du déclaratif employeur (déclaration annuelle + Document Unique d'Evaluation des Risques) : informations sur les risques présents dans l'entreprise
- Du déclaratif salarié (questionnaire de pré-visite connectée : les questions portent sur l'exposition aux risques professionnels et l'état de santé du salarié
- Du suivi médical
- De données de terrain (AMT)

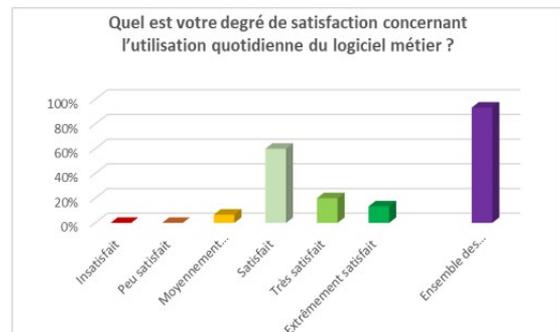

*Figure 4 Perception par les professionnels du SPSTI de l'utilisation du logiciel métier*

### 4.2 Une transformation organisationnelle concomitante à l'introduction de la data

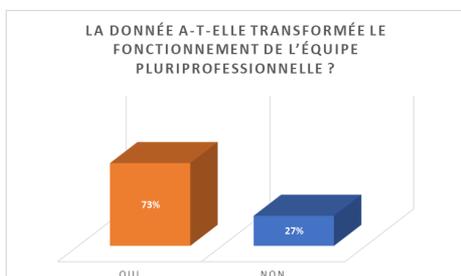

*Figure 5 Perception de l'impact de la data sur le fonctionnement de l'équipe pluriprofessionelle*

Il émerge des résultats que les professionnels du SPSTI, quel que soit leur statut professionnel, considèrent que l'introduction de l'utilisation de la donnée dans leur activité quotidienne a transformé le fonctionnement de l'équipe pluriprofessionnelle (Figure 5). Les demandes font l'objet d'échanges en réunion d'équipe et permettent de véritables débats intermétiers (Clot, 2008). Une base commune de partage existe autour de laquelle chacun peut apporter son éclairage. L'utilisation de ces données en équipe permet d'objectiver les situations et de réfléchir sur les possibilités d'actions. Lors d'une demande sur un poste, il apparaît qu'un autre poste de travail occupé par les mêmes salariés émerge comme particulièrement pathogène. Des actions conjointes (Visites médicales avec un focus sur ces questions, recherche bibliographique, statistique sur les visites à la demande et inaptitudes…) permettent de préparer le pré-diagnostic et la méthodologie à proposer à l'employeur. L'objectivation des données de santé et de la perception des travailleurs autorisent de renégocier la demande afin d'obtenir l'adhésion de l'employeur et des travailleurs. La data, intégrée dans un processus itératif fondé comme une expérience de vie (Varela et al., 1993), devient un objet intermédiaire (Vinck, 2009) qui ouvre également des possibilités de mise en relation de mondes exogènes, (travail réel, experts, employeur, IRP) dans l'objectif d'une construction commune.

L'acceptabilité de ces outils a été anticipée lors de leur mise en place. Lors de la conception de la plateforme numérique et des outils métier, un travail d'analyse et de co-construction a été mené de front afin d'intégrer les professionnels du SPSTI dans un environnement où l'outil participerait de la capabilité. Il a été possible de montrer que les asymétries statutaires effectives ne condamnaient pas les personnes à des actions prédéfinies, mais qu'en situation, elles pouvaient se saisir de sujet et réaliser des actions de tous type (Gouvenelle et al., 2022). Les transformations organisationnelles ont permis de développer le travail collectif au sein de l'équipe pluriprofessionnelle. Les résultats illustrés à la Figure 6 montrent que les différents statuts professionnels participent à la construction du diagnostic santé travail. Le médecin du travail demeure en position dominante dans l'équipe, en tant que coordinateur. Cependant, on constate une forme de décloisonnement entre les champs professionnels « médical », « prévention » et « administratif » avec des actions et objectifs communs pour répondre aux besoins des adhérents. Ce sont des dynamiques favorables à la construction d'un fonctionnement pluridisciplinaire des équipes pluriprofessionnelles. L'évolution du cadre participatif, du positionnement des métiers les plus invisibilisés (assistant, infirmier) a également été un levier pour proposer cette organisation au sein de laquelle, si les asymétries sociales et professionnelles n'ont pas disparues, elles apparaissent comme lissées (Gouvenelle, 2021). Le travail en équipe pluriprofessionnelle a donc évolué à la fois du fait des changements organisationnels et de l'apparition et l'utilisation de la donnée. Il reste cependant des disparités selon les équipes, avec des possibilités hétérogènes d'échanges, de co-construction et de travail collectif.

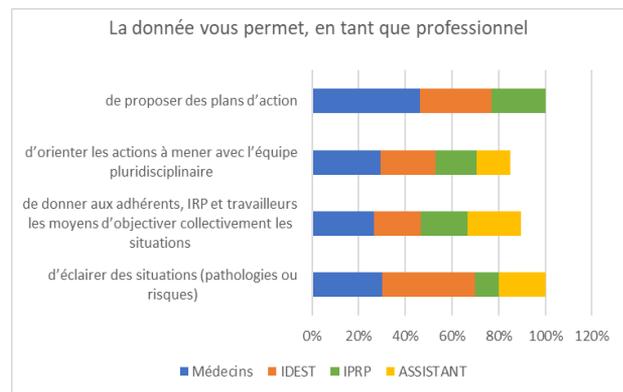

*Figure 6 L'utilisation de la donnée par les professionnels du SPSTI*

Un dernier élément à prendre en compte est le retour unanime d'un manque de formation qui participe à un sentiment de charge de travail augmentée et à une forme de *« frustration »*. Cela s'explique par l'aspect chronophage du temps de récolte et de traitement de données pour plusieurs raisons. Les personnes ont des difficultés à déterminer les objectifs et les caractéristiques des données en fonction de ce qu'ils cherchent à faire. Ensuite, ils ont des difficultés à manipuler les outils, à faire des tris, à manier les données, les échantillons et peuvent se sentir perdus aux milieux d'une masse de données à disposition sans savoir où aller, ce qui peut entraîner parfois l'abandon de la recherche initiée.

## 5   CONCLUSION

Les résultats soulignent que l'introduction de l'outil et la mise à disposition de la Data ne constitue pas en soi des leviers suffisants au développement de la pluridisciplinarité comme de la performance. Les outils sont perçus comme une aide mais également comme une charge de travail supplémentaire. La formation, l'appropriation des outils reste donc prioritaires. Cela autonomiserait les équipes pour construire des projets transversaux ou annuels mieux adaptés à leur effectif. L'approche développementale des organisations, des apprentissages ou de la conception des outils est donc centrale dans le processus. C'est le couplage de l'approche instrumentale et de l'activité, notamment les modalités de synchronisation de l'activité des personnes, cognitives comme opérateurs, qui sont à l'origine des dynamiques collectives (Rabardel, 1995). Elles contribueront aussi, par le développement de la pluridisciplinarité au sein des équipes pluriprofessionnelles, à des possibilités de renforcer le langage ou les pratiques communs aux métiers. De nouvelles constructions de normes propres à chaque métier sont également indispensables, notamment pour homogénéiser la construction des données dans le SPSTI. Cet objectif a pour ambition, par la poursuite des ajustements du cadre participatif, le débat inter et intra métier de construire un collectif de travail fondé sur une norme de métier « équipe pluriprofessionnelle » élaborée collectivement.

# 6 BIBLIOGRAPHIE

Barcellini, F., Détienne, F., & Burkhardt, J.-M. (2014). A situated approach of roles and participation in open source software communities. *Human-Computer Interaction*, *29*(3), 205-255.

Barlet, B. (2015). *De la médecine du travail à la santé au travail Les groupes professionnels à l'épreuve de la pluridisciplinarité*. Thèse de Doctorat en Sociologie, Université Paris-Ouest Nanterre.

Bourdieu, P. (1979). *La distinction. Critique du jugement social*. Editions de minuit.

Caroly, S. (2013). Les conditions pour mobiliser les acteurs de la prévention des TMS : construire du collectif de travail entre pairs pour développer le métier et favoriser le travail collectif pluri-professionnel. Le cas des médecins du travail. *Perspectives interdisciplinaires sur le travail et la santé*, *15*(2).

Caroly, S., Cholez, C., Landry, A., Davezies, P., Poussin, N., Bellemare, M., Coutarel, F., Garrigou, A., Chassaing, K., Petit, J., Baril-Gingras, G., Prudhomme, D., & Parrel, P. (2011). *Les activités des médecins du travail dans la prévention des TMS : ressources et contraintes*. ANR.

Caroly, S., & Sallah, M. (2015). *Tâches , objectifs des IPRP de SST et pluridisciplinarité. Analyse des résultats du questionnaire AFISST-PACTE des IPRP 2014*.

Clot, Y. (2008). *Travail et pouvoir d'agir*. PUF.

Coutarel, F. (2013). Le corps à l'effort : Un instrument du pouvoir d'agir. In *G. Boëtsch, D. Chevé & P. Blanchard, Corps et Couleurs* (p. 108-111).

Garrigou, A., Goutille, F., Gouvenelle, C., Vonarx, J., Lamarque, V., Swierczynski, G., Bussy, É., Canal-Raffin, M., Nascimento, A., Villa, A., Esterre, C., Aldana, J., Montigny, M., & Jolly, C. (2022, juillet 6). L'ergotoxicologie en actions – Mobiliser l'analyse de l'activité pour réduire les risques chimiques. *Actes du 56ème Congrès de la SELF, Vulnérabilités et risques émergents : penser et agir ensemble pour transformer durablement*.

Gouvenelle, C. (2021). *Les dynamiques des interactions asymétriques dans des collectifs pluridisciplinaires en SSTI. Une approche anthropologique et ergonomique de quatre projets de conception d'actions de prévention en santé au travail*. Thèse de doctorat STAPS, Anthropologie et Ergonomie, Université Clermont Auvergne.

Gouvenelle, C., Maudhuy, F., & Thorin, F. (2022). La data, un facteur de construction de la pluridisciplinarité. Freins et leviers de performance issus de la révolution numérique en Service de Prévention en Santé au Travail. *56ème Congrès de la SELF, Vulnérabilités et risques émergents : penseret agir ensemble pour transformer durablement. Genève, 6 au 8 juillet2022*, 204-209. www.ergonomie-self.org

Nielsen, J. (1994). *Usability Engineering*. Morgan Kaufmann.

Rabardel, P. (1995). *Les hommes et les technologies ; approche cognitive des instruments contemporains*. Armand Colin.

Récopé, M., Rix-Lièvre, G., Fache, H., & Boyer, S. (2013). La sensibilité à, organisatrice de l'expérience vécue. In *L. Albarello; J.-M. Barbier; E. Bourgeois & M. Durand. Expérience, activité, apprentissage* (p. 111-133). PUF.

Varela, F. J., Thompson, E., & Rosch, E. (1993). *L'inscription corporelle de l'esprit : Sciences cognitives et expérience humaine.* Seuil.

Vinck, D. (2009). De l'objet frontière à l'objet intermédiaire. Vers la prise en compte du travail d'équipement. *Revue d'anthropologie des connaissances*, *3*(1), 51-72.